%% file: acl2023_NESTA_arxiv.tex
\pdfoutput=1
\documentclass[11pt]{article}

\usepackage{ACL2023}

\usepackage{times}
\usepackage{latexsym}

\usepackage[T1]{fontenc}
\usepackage[utf8]{inputenc}

\usepackage{microtype}

\usepackage{inconsolata}

\usepackage{algorithm}
\usepackage{algorithmic}
\usepackage{tabularx}
\usepackage{multirow}
\usepackage{float}
\usepackage{graphicx}
\usepackage{multirow}
\usepackage{booktabs}
\usepackage{xcolor}
\usepackage{amsthm}
\usepackage{amsmath}
\usepackage{amssymb}
\usepackage{spverbatim}
\usepackage{newfloat}
\usepackage{listings}

\title{Learning Symbolic Rules over Abstract Meaning \\Representations for Textual Reinforcement Learning}

\author{ Subhajit Chaudhury, Sarathkrishna Swaminathan, Daiki Kimura, \\ {\bf Prithviraj Sen, Keerthiram Murugesan, Rosario Uceda-Sosa, Michiaki Tatsubori, } \\ {\bf Achille Fokoue, Pavan Kapanipathi, Asim Munawar and Alexander Gray}  \\ 
\{subhajit, sarath.swaminathan, keerthiram.murugesan, asim, alexander.gray\}@ibm.com\\
\{daiki, mich\}@jp.ibm.com, \{rosariou, achille, kapanipa\}@us.ibm.com\\
\\IBM Research}

\begin{document}
\maketitle
\begin{abstract}
Text-based reinforcement learning agents have predominantly been neural network-based models with embeddings-based representation, learning uninterpretable policies that often do not generalize well to unseen games. On the other hand, neuro-symbolic methods, specifically those that leverage an intermediate formal representation, are gaining significant attention in language understanding tasks. This is because of their advantages ranging from inherent interpretability, the lesser requirement of training data, and being generalizable in scenarios with unseen data. Therefore, in this paper, we propose a modular, NEuro-Symbolic Textual Agent (NESTA) that combines a generic semantic parser with a rule induction system to learn abstract interpretable rules as policies. Our experiments on established text-based game benchmarks show that the proposed NESTA method outperforms deep reinforcement learning-based techniques by achieving better generalization to unseen test games and learning from fewer training interactions.\footnote{Code available at \url{https://github.com/IBM/loa}}

%Results on text world commonsense show that NESTA outperforms deep reinforcement learning-based methods, specifically generalizing on unseen test games.
%Our model learns first-order horn clauses over symbolic facts extracted from Abstract Meaning Representations (AMR) of the observed text. We employ a differentiable inductive logic programming scheme to learn the lifted rules for the action commands that maximize the expected future reward. 
% The first-order abstraction of rule learning makes the symbolic policies generalizable to unseen games and interpretable, making them easier to debug. 
%Furthermore, our qualitative analysis shows that humans can easily change the learned rules to improve the generalization performance of unseen games.
\end{abstract} 

\input{sections/introduction_second.tex}
\input{sections/methods.tex}

\input{sections/experiments}
\input{sections/related_works}

\input{sections/conclusion.tex}

\section{Limitations}
% The neuro-symbolic rule learning presented in the paper can handle most generic text-based games. Only in a few specific use cases, additional training of the AMR parser would be required. Since AMR is used for symbolic representation for text-based games, the vocabulary of the extracted triples is limited by the vocabulary of PropBank semantic roles. For applications in a very specific kind of domain where the predicates and entities do not match with this pre-defined vocabulary (for example, specific financial, legal domains, etc.), the AMR semantic parsing engine needs to be retrained first on such specific data before using it for rule learning.

The neuro-symbolic rule learning presented in the paper can handle most generic text-based games. Only in a few specific use cases, additional training of the AMR parser would be required. Since AMR is used for symbolic representation for text-based games, the vocabulary of the extracted triples is limited by the vocabulary of PropBank semantic roles. For applications in a very specific kind of domain where the predicates and entities do not match with this pre-defined vocabulary (for example, specific financial, legal domains, etc.), the AMR semantic parsing engine needs to be retrained first on such specific data before using it for rule learning. However, even in the cases where the testing environment requires additional rules, NESTA allows human-in-the-loop debugging to conveniently add them making it adaptable to generic environments.

\section{Ethics Statement}
Our method uses a constrained set of action samples to generate the textual actions in each step. Since this action set is generated from a controlled vocabulary of actions and entities, the produced actions cannot contain harmful content like hate speech and racial biases. Furthermore, our neuro-symbolic model produces human interpretable rules for the action policy thereby making the model transparent and easier to control. Due to these reasons, the ethical risk from this work is low.

\bibliography{custom}
\bibliographystyle{acl_natbib}

\end{document}

%% file: sections/introduction_second.tex
\section{Introduction}

 \begin{figure}[tb]
		\centering
        \includegraphics[width=0.96\linewidth]{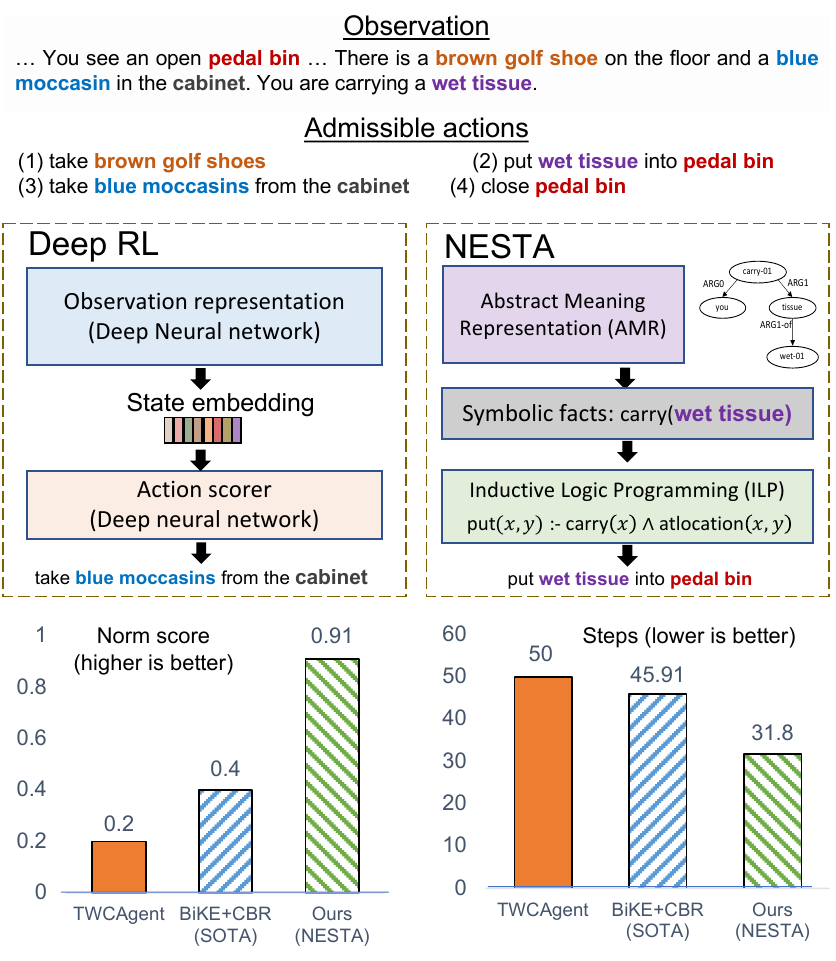}
		\caption{Our proposed NESTA model learns interpretable action rules as policy using ILP and outperforms SOTA deep RL methods on TBGs.}
		\label{fig:front_page}
		\vspace{-3mm}
\end{figure}

%Text-based games are complex, interactive simulations in which text describes the game state and players make progress by entering text commands. They are fertile ground for language-focused machine learning research. In addition to language understanding, successful play requires skills like long-term memory and planning, exploration (trial and error), and common sense. (from cote2018)

  Text-based games (TBGs)~\cite{cote2018textworld} serve as popular sandbox environments for evaluating natural language-based reinforcement learning. The agent observes the state of the game in pure text and issues a textual command to interact with the environment. TBGs are partially observable where the full state of the world is hidden and action commands facilitate the agent to explore the unobserved parts of the environment. The reward signal from the environment is used to improve the agent's policy and make progress in the game.
 
  Text-based games sit at the intersection of two research areas, i.e., language understanding and reinforcement learning. Existing RL agents for TBGs primarily use embeddings for observation as representations and are fed to an action scorer for predicting the next action~\cite{narasimhan-etal-2015-language, yuan2019counting, he-etal-2016-deep-reinforcement}, ignoring the advances in language understanding. On the other hand, there has been a recent surge in neuro-symbolic techniques, particularly those that use symbolic representations, for better language understanding~\cite{lu2021neurologic, kapanipathi2021leveraging} through reasoning. In light of exploring such advances for text-based reinforcement learning, this work proposes a neuro-symbolic approach. Our approach, named NESTA (NEuro Symbolic Textual Agent) is a modular approach comprising a generic semantic parser in combination with a symbolic rule induction system as shown in Figure~\ref{fig:front_page}. The semantic parser translates text into the form of symbolic triples. NESTA uses Abstract Meaning Representation~\cite{banarescu2013abstract} as the initial parse which is then transformed into triples. This symbolic representation is used by an adaptation of the Inductive Logic Programming (ILP) system using Logical Neural Networks~\cite{riegel2020logical} for learning horn clauses as action rules. 

  % LNN-ILP~\cite{sen2022neuro}
 
 NESTA, in comparison to other end-to-end learning approaches, has the following advantages: (a) modular language understanding using pre-trained large language models enabling our system to leverage the advances in semantic parsing. 
%For instance, in recent years, AMR parsers have gained significant improvements in Smatch score~\cite{cai2013smatch} using large language models~\cite{bevilacqua2021one, zhou2021amr}. 
While such modular semantic parsing-based techniques have been around for other NLP tasks such as reading comprehension~\cite{mitra2016addressing,galitsky2020employing}, knowledge base question answering~\cite{kapanipathi2021leveraging}, and natural language inference~\cite{lien2015semantic}, this work is the first to demonstrate the application for TBGs ; (b) learning symbolic rules for model-free RL using a neuro-symbolic framework facilitates inherent interpretability and generalizability to unseen situations~\cite{ma2021learning, jiang2019neural, dong2019neural}. The rules learned by NESTA are abstract and not specific to entities in the training data. These abstract action rules in policies for TBGs enable reasoning over unseen entities during training.     

Our main contributions in this work are: (1) We propose a novel and modular neuro-symbolic agent named NESTA. To the best of our knowledge, NESTA is the first to use a generic semantic parser with a rule learning system for TBGs, (2) Our empirical analysis of commonsense-aware textworld games shows that NESTA outperforms deep RL methods by a significant margin. We also show that NESTA has better sample efficiency compared to traditional text-based RL agents obtaining better test performance with up to $5\times$ lesser training interactions, and (3) Our method produces interpretable abstract rules from the rule induction system.

\begin{figure*}[tb]
		\centering
        \includegraphics[width=0.99\linewidth]{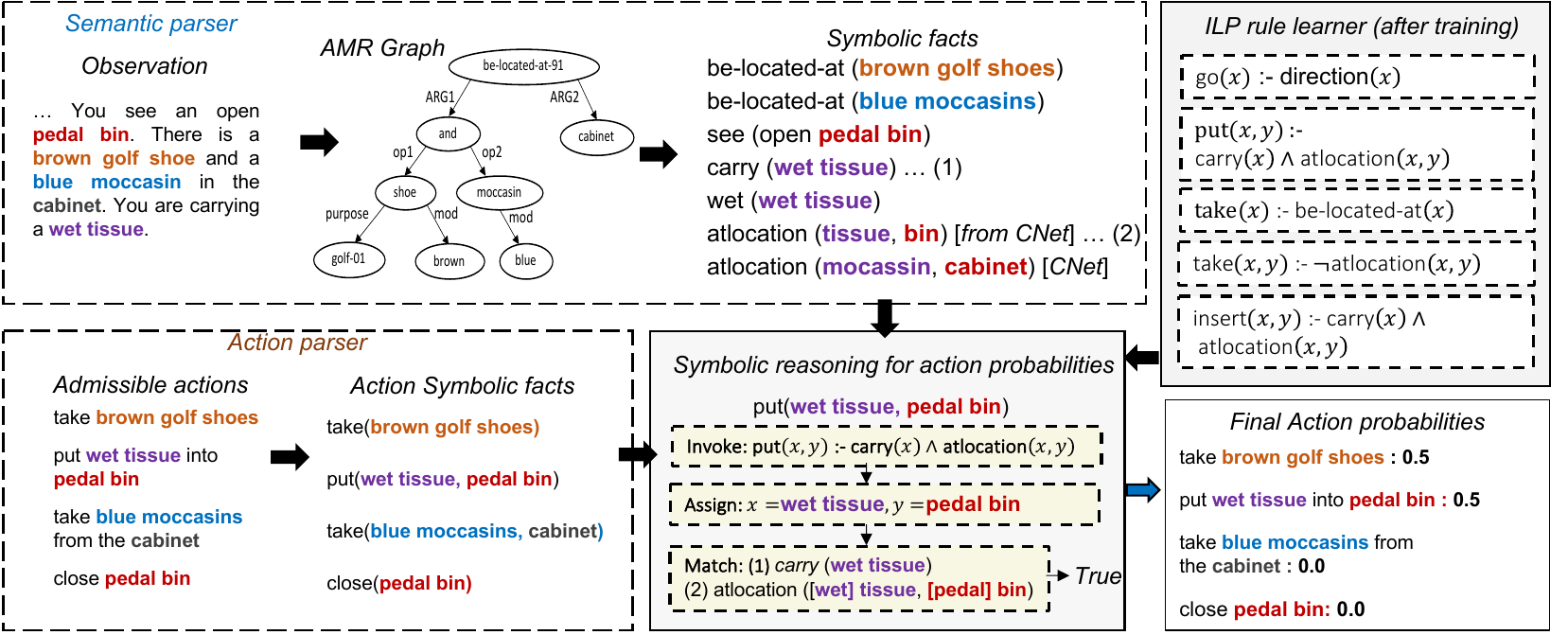}
		\caption{Overview of our method for neuro-symbolic reasoning. Our methods first extract symbolic facts from the surface text observations. During training, NESTA learns first-order lifted rules based on the reward signals. The learned rules are then used for obtaining the action probabilities.}
		\label{fig:method}
		\vspace{-3mm}
\end{figure*}

%% file: sections/methods.tex
\section{NEuro-Symbolic Textual Agent}
Text-based RL agents for TBGs interact with the environment using  text-only action commands and obtain feedback solely as textual observations. As the agent does not have access to global state information, it is modeled as a Partially Observable Markov Decision Process (POMDP)~\cite{kaelbling1998planning} represented as $(\mathcal{S}, \mathcal{A},  \mathcal{T},  R, \Omega, \mathcal{O})$, where $(\mathcal{S}, \mathcal{A},  \mathcal{T},  R)$ represent a Markov Decision Process. $\Omega$ represents the finite set of all observations, and $\mathcal{O}$ represents the observation function representing the conditional distribution over observations for a given action and next state. The goal of the agent is to learn optimal action probabilities at each step such that the expected future reward is maximized.

%Text-based deep reinforcement learning approaches~\cite{narasimhan-etal-2015-language,adolphs2019ledeepchef, ammanabrolu2020graph} use black-box neural network-based policies that overfit the training data and do not generalize to different configurations and unseen entities in test games. To alleviate this issue, we present 

We present 
\textbf{NE}uro-\textbf{S}ymbolic \textbf{T}extual \textbf{A}gent (\textbf{NESTA}), a modular approach for TBGs. Figure~\ref{fig:front_page} illustrates the overview of NESTA which comprises of  three primary components:  
% NESTA comprises three main components: 
(a) \textit{Semantic Parser}, which extracts symbolic representation of the text using  AMR as the generic semantic representation, (b) \textit{Rule Learner}, an ILP-based rule induction module, which learns logical rules that abstract out the entities in the games, making these rules generally applicable to test games containing unseen entities, and (c) \textit{Pruner}, that reduces the amount of branching factor at each step by pruning actions that do not contribute to the expected future reward. Below, we describe these components in detail.

\subsection{\textbf{Semantic Parser:} Text to symbolic triples using AMR}
The first step in NESTA is to translate the text into symbolic representation. To this end, inspired by works that address different NLP tasks~\cite{kapanipathi2021leveraging,galitsky2020employing,mitra2016addressing}, we use an AMR parser as a generic semantic parser. The use of a generic semantic parse such as AMR allows the system to benefit from independent advances in AMR research. For example, the performance of AMR has improved in Smatch score~\cite{cai2013smatch} from 70.9~\cite{van2017neural} to 86.7~\cite{lee-etal-2022-maximum} on LDC2017T10 in the last few years due to advances in large language models. The AMRs are subsequently transformed into a symbolic form using a deterministic AMR-to-triples approach. 
% We first extract the symbolic fact representation from the surface text using intermediate AMR semantic representation.
% Comments from daiki 
% 1. the observation text in Fig2 is not related to AMR graph.
% 2. The be-located-at() facts in Observation Symbolic facts in Fig2 must have arg2
% 3. \cite{zhou2021amr} is from https://github.com/IBM/transition-amr-parser/tree/83d67d54faaa66fb2e22178b4a54c6a1d64b102f which is version 0.4.2

\textbf{Abstract Meaning Representation (AMR)}:
AMR parsing produces rooted, directed acyclic graphs from the input sentences, where each node represents concepts from propbank frames~\cite{kingsbury2002treebank} or entities from the text. The edges represent the arguments for the semantic frames. Fig.~\ref{fig:method} shows the AMR graph generated from the sentence ``\textit{There is a brown golf shoe and a blue moccasin on the cabinet.}''. The resultant AMR graph is rooted at the propbank frame \texttt{be-located-at-91} with ARG1 and ARG2 edges leading to its children. The other parts of the graph are used to describe the entities for ``brown golf shoe'' and ``blue moccasin''. We use StructBART~\cite{zhou2021amr} for parsing a text to AMR. 
%For explicitly aligning the target nodes to sentence input, the model uses an action-pointer transition system which combines the advantages of both the transition-based approaches and general graph-generation approaches. The model has the transition as well as the pointer mechanism through straightforward modifications within a single Transformer architecture. Previous methods have used Stanford OpenIE~\cite{angeli2015leveraging} or regular expression-based predicate extraction for obtaining semantic representations. However, while AMR aligns the graph nodes with semantic frames from propbank, OpenIE and regex use surface form for entity-relation extraction without a well-defined predicate argument definition. Therefore, AMR parsing lends naturally to the next step of symbolic fact extraction.

\textbf{AMR-to-triples}: 
We design an AMR-to-triples module to extract a set of symbolic facts consisting of generic domain-agnostic predicates from the AMR semantic representation. Fig.~\ref{fig:method} shows the extraction of facts from AMR. The AMR-to-triples module performs a set of graph operations to extract propbank nodes as the predicates and the children entities as the arguments. In the example, the two operands of the ``and'' node are converted into two symbolic facts with \texttt{be-located-at} predicate with two separate entities of ``brown golf shoe'' and ``blue moccasins''. We convert the symbolic facts to unary predicates. For example, we convert the \texttt{be-located-at} predicate with two arguments into single argument facts. 
% Similarly, we remove ARG0 from the predicates because usually, that represents the agent. 
These simplifications result in some loss of representational power but make the task of rule learning simpler. 
% These operations are performed over the domain-independent AMR graphs and therefore can be applied for symbolic predicate extraction for most generic text-based environments. 
We also add the commonsense predicates from conceptnet subgraph~\cite{speer2017conceptnet} provided by the TWC environment~\cite{murugesan2020textbased}.

\subsection{Rule Learner: ILP from Rewards}
In order to learn interpretable rules that can be debugged by humans, we use the symbolic representation obtained from the above step. 
% Example of rules learned in this step can be $\textnormal{put}(x,y):-\; \textnormal{carry}(x) \wedge \textnormal{atlocation}(x,y)$, . 
Such symbolic rules are learned from reward signals by interacting with the environment. For this purpose, we use Inductive Logic Programming (ILP) in an RL setting with the objective of expected future reward maximization. We use Logical Neural Networks (LNN) as the differentiable rule learning engine.

% Reinforcement learning methods~\cite{williams1992simple} optimize policy for . Therefore, we use Logical Neural Networks (LNN) to learn rules using gradient-based training.

\textbf{Logical Neural Networks}: 
LNN~\cite{riegel2020logical} proposes a differentiable rule learning framework that retains the benefits of both neural networks and symbolic learners. It proposes a logical neuron that has the core properties of gradient-based learning similar to a standard neuron but adds logic-aware forward functions and constrained optimization making it suitable for logical operations. This can be illustrated on 2-input logical conjunction (AND) neuron with $(x, y)$ as two logical inputs to the conjunction node. The LNN conjunction neuron generalizes the classical AND logic for real-valued logic by defining a noise threshold ($\alpha$). The real-values in $[\alpha, 1]$ and $[0, 1- \alpha]$ signify a logical \textit{high} and logical \textit{low} respectively. To emulate an AND neuron, LNN uses the standard truth table of the conjunction (AND) gate to obtain the following constraints, 
\begin{equation*}
	\begin{split}
		f(x, y) & \leq 1 - \alpha, \quad  \forall  x, y \in [0, 1-\alpha] \\
		f(x, y) & \leq 1 - \alpha, \quad  \forall  x \in [0, 1-\alpha], y \in [\alpha, 1] \\
		f(x, y) & \leq 1 - \alpha, \quad  \forall  x \in [\alpha, 1], y \in [0, 1-\alpha] \\
		f(x,y) & \geq \alpha, \quad \quad \;\;\; \forall  x,y\in [\alpha, 1]
	\end{split}.
\end{equation*}
LNN uses the forward function as the weighted \L ukasiewicz t-norm, $f(x, y; \beta, w_1, w_2) = \beta - w_1 ( 1 - x) - w_2 (1 - y)$, where $\beta, w_1, w_2$ are the bias and weights of the inputs. Given a target label, the weights and biases are tuned to learn the logical rule that best describes the data.

\textbf{ILP-based reward maximization}: Our ILP rule learner is based on the LNN rule learning implementation in \citet{sen2022neuro}. However, our rule-learning model makes significant modifications to adapt the previous algorithm for model-free policy optimization suitable for text-based RL. Consider the state transition at time step $t$ as $(o_t, a_t, r_t, o_{t+1})$, where $o_t$ represents the textual observation, $a_t$ is the action command that yields the reward $r_t$ and takes the agent to the next state with observation $o_{t+1}$. AMR-to-triples semantic parser is used to obtain the symbolic state $s_t$ (list of symbolic facts) from $o_t$ as shown in Figure~\ref{fig:method}. At each step, the agent has to choose from a set of admissible action commands which are also converted to their symbolic form. Starting from an initial random policy $\pi$, we sample trajectories $\tau \sim \pi$ and store the transitions $(s_t, a_t, r_t, s_{t+1})$ in a buffer $\mathcal{B}$. We also store the admissible actions set $\text{adm}_t$ and the discounted future reward $g_t = \sum_{k=t}^{T}\gamma^{k-t}r_k$, for each step in the buffer, where $\gamma$ is the discount factor. 
%%%%%%%%%%%%%
% In distribution
%%%%%%%%%%%%%

\begin{table*}[htb]
\centering
\resizebox{0.92\textwidth}{!}{%
\begin{tabular}{c|cc|cc|cc}
\hline
& \multicolumn{2}{c|}{\textbf{Easy}}          & \multicolumn{2}{c|}{\textbf{Medium}}                              & \multicolumn{2}{c}{\textbf{Hard}}           \\ \hline
\textbf{Methods} & \textbf{Steps}            & \textbf{Norm. Score}     & \textbf{Steps}            & \textbf{Norm. Score}     & \textbf{Steps}            & \textbf{Norm. Score}     \\ \hline

\textbf{Text} & 23.83 $\pm$ 2.16 & 0.88 $\pm$ 0.04 & 44.08 $\pm$ 0.93 & 0.60 $\pm$ 0.02 & 49.84 $\pm$ 0.38 & 0.30 $\pm$ 0.02 \\
\textbf{Text+CS} & 20.59 $\pm$ 5.01 & 0.89 $\pm$ 0.06 & 42.61 $\pm$ 0.65 & 0.62 $\pm$ 0.03 & 48.45 $\pm$ 1.13 & 0.32 $\pm$ 0.04 \\
\textbf{KG-A2C} & 22.10 $\pm$ 2.91 & 0.86 $\pm$ 0.06 & 41.61 $\pm$ 0.37 & 0.62 $\pm$ 0.03 & 48.00 $\pm$ 0.61 & 0.32 $\pm$ 0.00 \\
\textbf{BiKE} & 18.27 $\pm$ 1.13 & 0.94 $\pm$ 0.02 & 39.34 $\pm$ 0.72 & 0.64 $\pm$ 0.02 & 47.19 $\pm$ 0.64 & 0.34 $\pm$ 0.02
\\
\textbf{BiKE+ CBR} & 15.72 $\pm$ 1.15 & 0.95 $\pm$ 0.04 & 35.24 $\pm$ 1.22 & 0.67 $\pm$ 0.03 & 45.21 $\pm$ 0.87 & 0.42 $\pm$ 0.04 \\

\hline

\textbf{NESTA} & \textbf{2.40 $\pm$ 0.00} & \textbf{1.00 $\pm$ 0.00} & 31.44 $\pm$ 2.08 & 0.80 $\pm$ 0.04 & 42.68 $\pm$ 6.01 & \textbf{0.85 $\pm$ 0.05} \\

\textbf{NESTA + OR} & 3.44 $\pm$ 2.08 & \textbf{1.00 $\pm$ 0.00} & \textbf{11.76 $\pm$ 1.78} & \textbf{0.98 $\pm$ 0.03} & \textbf{35.84 $\pm$ 7.88} & \textbf{0.85 $\pm$ 0.09} \\
\hline
\hline
\textbf{Human} & 2.12 $\pm$ 0.00 & 1.00 $\pm$ 0.00 & 5.33 $\pm$ 0.00 & 1.00 $\pm$ 0.00 & 15.00 $\pm$ 0.00 & 1.00 $\pm$ 0.00 \\
\hline
\end{tabular}}
\caption{Proposed NESTA model shows better performance in terms of normalized score and steps to reach the goal compared to Deep RL methods on unseen TWC \emph{in-distribution} games.}
\label{tab:twc_in}
\end{table*}

From the buffer $\mathcal{B}$, we find a set of template predicates $\mathcal{P}=\{ p \mid p \in s_t, \;\textnormal{  for  } s_t \in \mathcal{B} \}$, where $p \in s_t$ operation states whether facts with predicate $p$ exist in the symbolic state $s_t$. We also obtain a set of action predicates $\mathcal{A}=\{ a \mid a \in \text{adm}_t, \textnormal{  for  } \text{adm}_t \in \mathcal{B} \}$ finding all action predicates in the admissible action set. 
We initialize ILP rule learner $\pi_a(\theta)$ for each action predicate $a \in \mathcal{A}$. Action predicates for TBGs typically coincide with the action verbs. The LNN policy is formulated as a weighted conjunction operation over the template predicates $\mathcal{P}$. The likelihood of action $a$ for abstract lifted variables $x,y$ is given as a conjunction template over the predicate list as follows: unary action likelihood is given as $L(a(x)|s_t) = \bigwedge_k w_k p_k(x)$ and binary action likelihood is formulated as $L(a(x,y)|s_t) = \bigwedge_k w_k p_k(x) \bigwedge_m w_m q_m(x,y)$. The predicates $p_k$ and $q_m$ are 1 and 2 arity predicates in $\mathcal{P}$ respectively and $\bigwedge$ represents the LNN's logical conjunction operator. The weights $w_k$ and $w_m$ constitute the LNN parameters $\theta$ that are updated during training. At any given step, the likelihood of each action is normalized over all actions in the admissible action set to obtain the action probabilities. 

For training the rule learning model $\pi_a(\theta)$ for a specific action $a$, we only extract transitions from the buffer containing the action $a$ and store it in a sub-buffer $\mathcal{B}_a$. The model is updated following the policy gradient loss, $\mathcal{L} = \nabla_{\theta} \mathbb{E}_{(s_t, g_t) \sim \mathcal{B}_a} \log(\pi_a(a_t=a|s_t)g_t $, where the trajectories are sampled from $\mathcal{B}_a$. We assume that $\pi_a$ gives normalized probabilities for this loss formulation. Therefore, this training procedure yields separate rules learned for each action predicate. Figure~\ref{fig:method} shows the learned rules for each action. 
% We use a weight threshold of 0.5 and only show predicates in the rule that have weights above the threshold. 

\textbf{Generalization under Distribution Shift}:
Having learned the action rules for each action predicate using dedicated ILP models, NESTA uses the rules for obtaining the action probabilities at each step. This process consists of three steps: (a) For each action in the admissible action list, \textbf{invoke} the learned rule for that action predicate, (b) \textbf{Assign} the abstract variables with the symbolic action arguments, and (c) \textbf{Match} the symbolic facts using entity alignment by root noun matching (instead of an exact match). The probabilities are then obtained by the LNN conjunction node feed-forward operation based on the current weights. This procedure is also used for sampling during training.

Figure~\ref{fig:method} shows the reasoning steps for fixed weights after training is complete. Since the rules learned by NESTA abstract out the entities in the form of lifted variables, human interpretability and generalization to unseen entities is a natural advantage of our method. In addition to this, since we modularize language understanding and RL policy learning into separate modules, our LNN symbolic learner can solely focus on optimal reward performance leading to sample-efficient learning.

\subsection{Pruner: Irrelevant Action Pruning by Look-Ahead}
The third module in NESTA, tackles the large action space problem in TBGs by removing actions from the admissible commands that do contribute to future rewards in the games. A large number of possible actions at each step can increase the branching factor of the agent at each step during the training and testing leading to a combinatorially large search problem. We employ a look-ahead strategy to find out which actions do not contribute to future reward accumulation. For example, the action  \texttt{examine}($x$) returns the description of the entity $x$, but does not change the state of the game and does not contribute to future rewards. However, for the action \texttt{take}($x$), although an immediate reward is not obtained on execution, it leads to a future reward when the object $x$ is put in the correct container $y$ using the \texttt{put}($x, y$) command. Therefore, the action command of type \texttt{examine}($x$) can be pruned but \texttt{take}($x$) is essential and hence cannot be pruned. This can be computed by looking ahead from the current step and comparing the future reward if that particular action was removed from the trajectory. 

Due to AMR error propagation and undesirable credit assignment (for example, \texttt{examine}($x$) command issued just before a rewarded action), the rule learner can assign high action probabilities to non-contributing actions. Therefore, the pruner module is desirable to remove such action predicates ($a$) by evaluating the total reward in action trajectories with and without the particular action predicate $a$. More specifically, for each episodic trajectory, we remove the action predicates $a$ and re-evaluate the episodic reward obtained from the environment. If the average episodic reward in both cases, with and without removal of $a$ is the same then the action predicate is not contributing to the future reward. Therefore, it can be removed from the original action predicate set, $\mathcal{A}$ to obtain the pruned action set $\mathcal{A}_\text{pruned}$ for which LNN models are learned.

\section{Outlier Rejection in Policy Training}

The training samples that NESTA collects from interacting with the environment can be noisy and this can affect learning a good policy. There exists two sources of noise: (a) \textbf{AMR noise}, where AMR incorrectly parses the surface text resulting in erroneous identification of entity extraction or relationships between entities, and (b) \textbf{RL credit assignment noise}, where discounted reward gives reward to a suboptimal action taken right before a correct action. Although symbolic reasoners have the advantages of learning from fewer data and better generalization, they are not robust to noise. We mitigate the effect of noise in LNN policy training by using a consensus-based noise rejection method. 

Our noise rejection method trains the LNN Policy on multiple subsets of training data and selects the model with the smallest training error as the best model. The multiple subsets of training data are prepared as follows - for each training subset, a particular predicate $p$ from the predicate list $\mathcal{P}$ is given priority. We only choose state transitions that contain the predicate $p$ ensuring that this predicate will be part of the final learned rule, thus eliminating the source of AMR noise for this predicate (such a subset is rejected if the number of such transitions is less than some threshold percentage). Subsequently, the resulting transitions are sorted by the discounted reward $g_t$ and we only retain the top first $k$\% of this sorted data as training data. This encourages action transition with more immediate average reward gains to constitute the training data. 

% Finally, the best rule is obtained from the LNN model with the lowest training loss among these training subsets.

% \begin{algorithm}
% \DontPrintSemicolon
  
%   % \KwInput{Data (observation, discounted reward}
%   % \KwOutput{LNN Policy}
%   % \KwData{Testing set $x$}
%   Data = (observation, label)
  
%   Best LNN Policy = None
  
%   Best\_Error = inf
  
%   \For{predicate \textbf{in} Data}
%   {
%     Sorted\_Data = sort(Data, keys=[predicate, label], ascending=False) \;
      
%     Train\_Data = First 70\% of Sorted\_Data \;
      
%     Test\_Data = Last 30\% of Sorted\_Data \;
      
%     Train\_Error = LNN\_Policy.train(Train\_Data) \;
      
%     \tcc{splitting data after sorting minimizes the presence of noise in train split}
      
%     \For{test\_data_i, test\_label_i\: \textbf{in}\: Test\_Data}
%     {
%       prediction_i = LNN\_Policy.predict(test\_data_i) \;

%       error_i = error( test\_label_i, prediction_i) \;
    
%       \If{$error_i \leq Train\_Error$}
%       { Train\_Data.append((test\_data_i, test\_label_i)) }
%     }
    
%     Train\_Error = LNN\_Policy.train(Train\_Data) \;
%     \tcp*{Train LNN Policy on updated Train\_Data}
    
%     \If{$Train\_Error \le Best\_Error$}
%     {
%       Best LNN Policy = LNN\_Policy
%       Best\_Error = Train\_Error
%     }

%   }
% \caption{Noise Rejection in Policy training}
% \label{algo:noise_rejection}
% \end{algorithm}

%% file: sections/experiments.tex
%%%%%%%%%%%%%
% Out of distribution
%%%%%%%%%%%%%

\begin{table*}[tb]
\centering
\resizebox{0.92\textwidth}{!}{%
\begin{tabular}{c|cc|cc|cc}
\hline
& \multicolumn{2}{c|}{\textbf{Easy}}          & \multicolumn{2}{c|}{\textbf{Medium}}                              & \multicolumn{2}{c}{\textbf{Hard}}           \\ \hline
\textbf{Methods} & \textbf{Steps}            & \textbf{Norm. Score}     & \textbf{Steps}            & \textbf{Norm. Score}     & \textbf{Steps}            & \textbf{Norm. Score}     \\ \hline

\textbf{Text} & 29.90 $\pm$ 2.92 & 0.78 $\pm$ 0.02 & 45.90 $\pm$ 0.22 & 0.55 $\pm$ 0.01 & 50.00 $\pm$ 0.00 & 0.20 $\pm$ 0.02 \\
% \textbf{DRRN} & 29.71 $\pm$ 1.81 & 0.76 $\pm$ 0.05 & 45.18 $\pm$ 1.19 & 0.56 $\pm$ 0.02 & 50.00 $\pm$ 0.00 & 0.21 $\pm$ 0.02 \\
\textbf{Text+CS} & 27.74 $\pm$ 4.46 & 0.78 $\pm$ 0.07 & 44.89 $\pm$ 1.52 & 0.58 $\pm$ 0.01 & 50.00 $\pm$ 0.00 & 0.19 $\pm$ 0.03 \\
\textbf{KG-A2C} & 28.34 $\pm$ 3.63 & 0.80 $\pm$ 0.07 & 43.05 $\pm$ 2.52 & 0.59 $\pm$ 0.01 & 50.00 $\pm$ 0.00 & 0.21 $\pm$ 0.00 \\
\textbf{BiKE} & 25.59 $\pm$ 1.92 & 0.83 $\pm$ 0.01 & 41.01 $\pm$ 1.61 & 0.61 $\pm$ 0.01 & 50.00 $\pm$ 0.00 & 0.23 $\pm$ 0.02 \\ 
\textbf{BiKE + CBR} & 17.15 $\pm$ 1.45 & 0.93 $\pm$ 0.03 & 35.45 $\pm$ 1.40 & 0.67 $\pm$ 0.03 & 45.91 $\pm$ 1.32 & 0.40 $\pm$ 0.03 \\
\hline
\textbf{NESTA} & \textbf{2.40 $\pm$ 0.00} & \textbf{1.00 $\pm$ 0.00} & 5.56 $\pm$ 0.53 & 1.00 $\pm$ 0.00 & 38.88 $\pm$ 3.24 & \textbf{0.94 $\pm$ 0.04} \\

\textbf{NESTA + OR} & 3.28 $\pm$ 1.76 & \textbf{1.00 $\pm$ 0.00} & \textbf{3.60 $\pm$ 0.00} & \textbf{1.00 $\pm$ 0.00} & \textbf{31.40 $\pm$ 6.38} & {0.91 $\pm$ 0.05} \\
\hline
\hline
\textbf{Human} & 2.24 $\pm$ 0.00 & 1.00 $\pm$ 0.00 & 4.40 $\pm$ 0.00 & 1.00 $\pm$ 0.00 & 17.67 $\pm$ 0.00 & 1.00 $\pm$ 0.00 \\
\hline
\end{tabular}}
\caption{Normalized score and number of steps to reach the final goal for various methods on unseen TWC \emph{out-of-distribution} games. NESTA shows large improvements over previous Deep RL methods, especially for \textit{hard} games. \textbf{OR} is Outlier Rejection. }
\label{tab:twc_out}
\end{table*}

%%%%%%%%%%%%%%%%%
% Sample efficiency  %
%%%%%%%%%%%%%%%%%

\section{Experimental results}

Our experiments are designed to answer these questions that analyze if NESTA can overcome the common drawbacks of deep RL methods: (i) Can NESTA enable better generalization in test environments? (ii) Does NESTA improve upon sample efficiency while still maintaining good reward performance, (iii) Are the rules learned by NESTA, human interpretable? For comparing the performance of various methods, we use the metrics of \textit{normalized score} (total reward from the games normalized by maximum reward) and \textit{number of steps} to reach the goal (lower is better). Our experiments were conducted on Ubuntu 18.04 operating system with NVidia V100 GPUs.

\subsection{Environment}
We use the textworld commonsense (TWC) environment~\cite{murugesan2020textbased} for empirical evaluation of our method. The goal here is to clean up a messy room by placing the objects in the correct containers. The game provides conceptnet sub-graphs relating the game entities which are used as commonsense graphs. TWC provides two splits of testing games: (i) \textit{in-distribution} games that have the same entities as training games but unseen object-container configuration, and (ii) \textit{out-of-distribution} games that use new objects not seen during training. This provides a systematic framework for measuring generalization in NESTA and other baseline agents for both within-training distribution and out-of-training distributions. Since we are focusing on generalization aspects, we do not use other textworld games~\cite{cote2018textworld, hausknecht2020interactive} because these environments primarily focus on the agent's exploration strategies and are therefore not suitable to evaluate the agent's generalization ability.

\subsection{Agents}
For baseline agents, we report performance by these deep RL-based methods: (1) \textbf{Text}-based agent that uses a GRU network for observation representation and action scorer units, (2) \textbf{TWCAgent (Text + CS)} that uses combined textual and commonsense embeddings for action scoring, (3) \textbf{KG-A2C}~\cite{ammanabrolu2020graph} that uses extracted knowledge graphs as input, (4) \textbf{BiKE}~\cite{murugesan2021efficient} which leverages graph structures in both textual and commonsense information and (5) \textbf{CBR}~\cite{atzeni2021case} which is the SOTA method using case-based reasoning for improving generalization in text-based agents. We did not compare with previous neuro-symbolic methods~\cite{kimura2021neuro, chaudhury2021neuro} because they use a hand-crafted game-specific predicate design scheme that was not available for TWC.

\subsection{Generalization to Test Games}
We evaluate the generalization ability of NESTA on TWC \textit{easy}, \textit{medium} and \textit{hard} games. 
% For all the games humans get a perfect score with the following average steps: (1) \textit{In-dist}: easy~(2.12), medium (5.33), hard (15.0), (1) \textit{Out-of-dist}: easy~(2.24), medium (4.4), hard (17.67). 
Table~\ref{tab:twc_in} and Table~\ref{tab:twc_out} shows the performance of baseline and our agents on \textit{in-distribution} and \textit{out-of-distribution} games, including the human performance from \citet{murugesan2020textbased}. For the baseline models, we report scores from ~\citet{atzeni2021case}. For NESTA, we report the mean of 5 independent runs. 

For easy games, NESTA gets a perfect score outperforming previous games with similar steps as human performance. For medium and hard games, NESTA greatly surpasses the SOTA agent and needs a lesser number of steps for both \textit{in-distribution} and \textit{out-of-distribution} games. For medium \textit{out-of-distribution} games, NESTA outperforms humans in terms of the number of steps. This might be due to the fact that during human annotation, the subjects would take a larger number of steps for the initial few games due to trial-and-error, thus increasing the average number of steps.

While easy and medium games have a single-room setting, hard games present a two-room setting where the agent might require picking up an object in room 1 and putting it in a container in room 2. This requires learning a complex strategy especially for generalizing to unseen entities. Our method NESTA scores significantly higher compared to SOTA on hard games, thus exhibiting the ability of our method to generalize in complex settings while deep RL methods fail to generalize due to overfitting the training data. Furthermore, our outlier rejection model helps improve the number of steps to reach the goal for both \textit{in-distribution} and \textit{out-of-distribution} games.

\subsection{Ablation Results with Action Pruning}
% TBGs typically suffer from large action spaces~\cite{zahavy-nips-2018, fulda2017can}. We use symbolic action pruning for NESTA to remove unwanted actions from the admissible actions list. 
To study the effect of our action pruning module on deep RL agents, we implemented action pruning on the publicly available TWCAgent code from \citet{murugesan2020textbased}.  We follow the exact same methodology for TWCAgent that we used for the NESTA agent. Using the look-ahead method, we obtain $\mathcal{A}_{\text{retain}}$, the list of action verbs to retain at a specific episode (episode num 10 for this result). For all subsequent training steps, only action verbs $a \in \mathcal{A}_{\text{retain}}$ were retained from the admissible list. We also follow the same strategy for the test games.

Table ~\ref{tab:ablation} shows the results for action pruning for both TWCAgent and NESTA. Firstly, even without action pruning, NESTA outperforms the TWCAgent with action pruning. NESTA+AP shows a higher gain in performance compared to NESTA only, whereas TWCAgent did not exhibit such large improvements. We found that even without AP, TWCAgent learns to avoid sub-optimal actions. However, it suffers from overfitting and hence cannot generalize to unseen configurations and entities.

%%%%%%%%%%%%%
% Ablation  %
%%%%%%%%%%%%%

\begin{table}[tb]
\centering
\resizebox{0.46\textwidth}{!}{%
\begin{tabular}{lcc}
\hline
% \multicolumn{1}{c|}{}              & \multicolumn{2}{c}{Hard} \\ \hline
\multicolumn{3}{c}{\textit{In-distribution}}                  \\ \hline
\multicolumn{1}{c|}{}              & \textbf{Steps}    & \textbf{Norm. Score}   \\ \hline
\multicolumn{1}{l|}{\textbf{TWCAgent}}      & 47.77 $\pm$ 1.50        & 0.49 $\pm$ 0.04            \\
\multicolumn{1}{l|}{\textbf{TWCAgent + AP}} & 47.14 $\pm$ 0.85        & 0.61 $\pm$ 0.03            \\
\multicolumn{1}{l|}{\textbf{NESTA}}         & 43.44 $\pm$ 4.67       & 0.77 $\pm$ 0.08             \\
\multicolumn{1}{l|}{\textbf{NESTA + AP}}    & \textbf{35.84 $\pm$ 7.88}  & \textbf{0.85 $\pm$ 0.09}            \\ \hline
\multicolumn{3}{c}{\textit{Out-of-distribution}}              \\ \hline
\multicolumn{1}{l|}{}              & \textbf{Steps}    & \textbf{Norm. Score}   \\ \hline
\multicolumn{1}{l|}{\textbf{TWCAgent}}      & 50.00 $\pm$ 0.00      & 0.21 $\pm$ 0.05     \\
\multicolumn{1}{l|}{\textbf{TWCAgent + AP}} & 50.00 $\pm$ 0.00      & 0.37 $\pm$ 0.02     \\
\multicolumn{1}{l|}{\textbf{NESTA}}         & 47.52 $\pm$ 2.34      & 0.60 $\pm$ 0.15     \\
\multicolumn{1}{l|}{\textbf{NESTA + AP}}    & \textbf{31.40 $\pm$ 6.38}      & \textbf{0.91 $\pm$ 0.05}     \\ \hline
\end{tabular}
}
\caption{Ablation study showing the effect of our proposed symbolic action pruning (\textbf{AP}) on NESTA and TWCAgent for \textit{hard} games. Proposed action pruning method shows better improvements on \textbf{NESTA} model when compared to improvements on TWCAgent.}
\label{tab:ablation}
\end{table}

\subsection{Human-in-the-loop Rule Debugging}
% Each step of the NESTA pipeline is human interpretable. 
NESTA enables the user to verify all the learned rules. It provides the facility to add new rules that might be missing or edit the rules if they are sub-optimal. 
The ability of human-in-the-loop debugging is what sets NESTA apart from other methods that tend to provide some level of explainability. Table~\ref{tab:human} shows the human-interpretable learned rules for a particular training on hard games. The rule for $\text{take}(x,y)$ can be identified as sub-optimal because it implies that the agent should take any object that is present in a container $y$ present in the current room. The human-corrected rule implies the agent should only ``take'' objects that are not in their assigned location according to conceptnet facts. The human-corrected rule perfectly solves the \textit{out-of-distribution} hard games in close to the optimal number of steps. This demonstrates that NESTA's human-in-the-loop rule debugging feature can be readily used to achieve favorable performance gains.

%%%%%%%%%%%%%%%%%
% Human debug  %
%%%%%%%%%%%%%%%%%

\begin{table}[tb]
\centering
\resizebox{0.42\textwidth}{!}{%
\begin{tabular}{l}
\hline
\textbf{Learned rules for \textit{hard} games by NESTA} \\
\hline
$\textnormal{go}(x):-\; \textnormal{direction}(x)$ \\
$\textnormal{take}(x):-\; \textnormal{be-located-at}(x)$ \\
$\textnormal{take}(x,y):-\; \textcolor{red}{\textnormal{be-located-at}(y)}$ \\
$\textnormal{put}(x,y):-\; \textnormal{carry}(x) \wedge \textnormal{atlocation}(x,y)$ \\
$\textnormal{insert}(x,y):-\; \textnormal{carry}(x) \wedge \textnormal{atlocation}(x,y)$ \\
\hline
In-distribution norm score: $0.71$ (Steps: $46.4$) \\
Out-distribution norm score: $0.85$ (Steps: $37.4$)\\
\hline
\textbf{After rule correction by human}\\
\hline
$\textnormal{take}(x,y):-\; \textcolor{violet}{\neg \textnormal{\textbf{atlocation}}(x,y)}$ \\
\hline
In-distribution norm score: \textbf{0.88} (Steps: \textbf{42.4}) \\
Out-distribution norm score: \textbf{1.0} (Steps: \textbf{19.8})\\
\hline
\hline
\end{tabular}
}
\caption{Action rules learned for NESTA (seed=2) on \textit{hard} games. The rules are human-interpretable making them easy to debug. We highlight that the rule learned for $\textnormal{take}(x,y)$ is sub-optimal and can be improved by human-in-the-loop correction of that single rule with large performance gains. }
\label{tab:human}
\end{table}

\subsection{Sample efficient learning}
\begin{figure}[tb]
		\centering
        \includegraphics[width=0.99\linewidth]{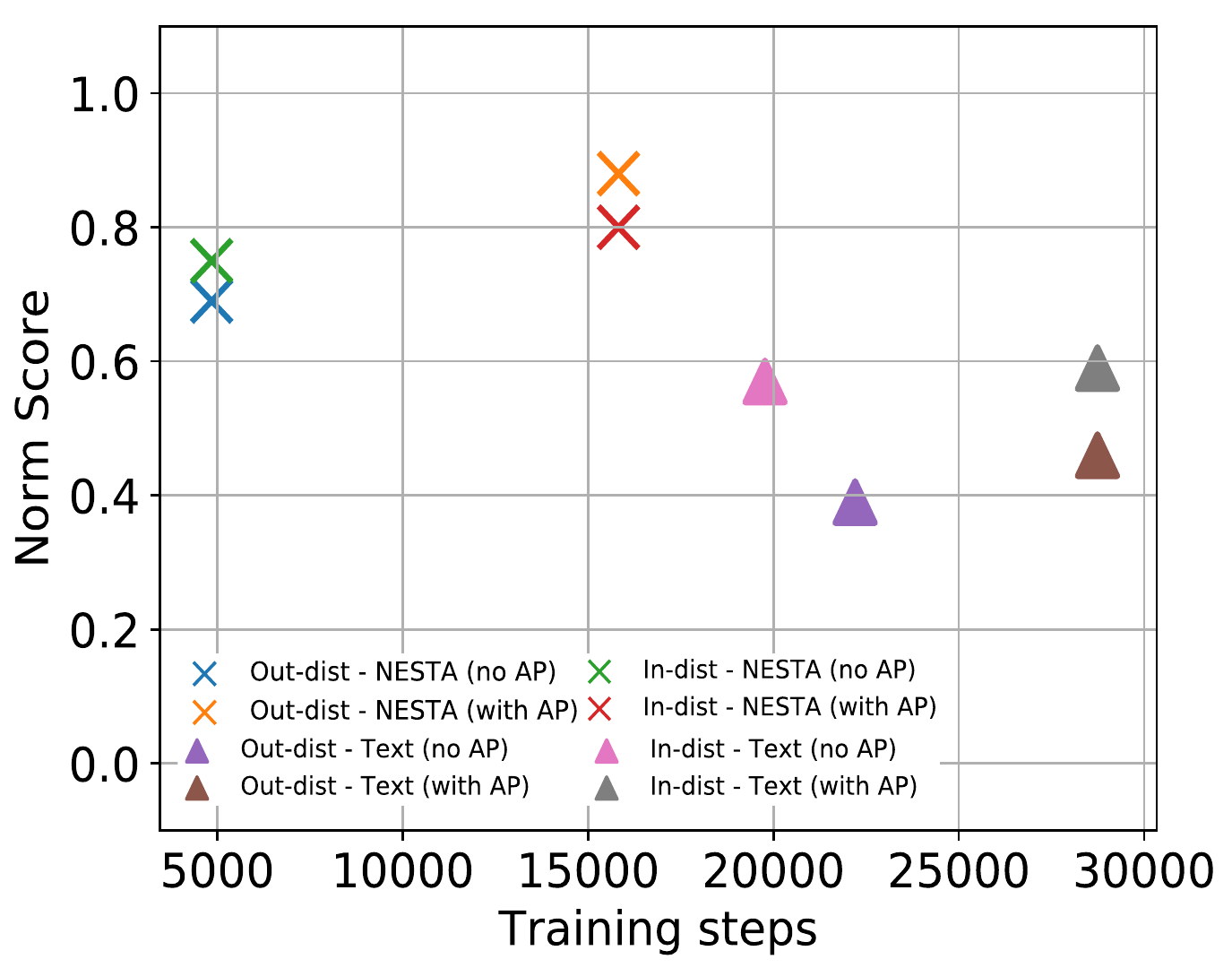}
	\caption{Normalized score obtained by NESTA and deep RL text agent on TWC games. NESTA achieves a higher score compared to the deep agent even when learning from 5x lesser training interactions. The top-left part represent better performance while the bottom-right part conveys worse performance.}
		\label{fig:sample_eff}
\end{figure}

% Deep RL methods suffer from poor sample efficiency requiring a large number of training interactions to obtain a good performance. 
We hypothesize that deep RL policies require a large number of training interactions because they learn both language understanding and action scoring from rewards ignoring external language pre-training. NESTA, on the other hand, decouples language understanding to AMR-based semantic representations while the LNN-ILP rule learner can focus on RL policy optimization resulting in learning from fewer samples. Figure~\ref{fig:sample_eff} shows that the NESTA model obtains better scores for both \textit{in-distribution} and \textit{out-distribution} games at much fewer training interactions compared to the deep RL text agent. In fact, NESTA can outperform text agents even when it learns from $5\times$ lesser training interactions.

We also computed computational time for NESTA compared to neural agents. Average computational times (out-of-distribution) required for each step for NESTA compared to neural agents. For easy games, the average computation time for neural agents was $0.12\pm0.06$ s, and that for NESTA was $0.16\pm0.05$. The corresponding numbers for medium games were $0.17\pm0.06$ and $0.22\pm0.06$ respectively. NESTA requires extra time due to parsing. However, since it has a lower overall number of steps (almost 5 times lower for easy/medium games from Table \ref{tab:twc_out}), time per game would be lower or comparable.

%% file: sections/related_works.tex
\section{Related Work}
\label{sec:related_work}

% In this section, we review some of the relevant works in text-based RL literature.

\textbf{Text-only Agents:}
Early work on text-based reinforcement learning agents used an LSTM-based representation learning from textual observations ~\cite{narasimhan2015language}, and Q-learning ~\cite{watkins1992q} in the action scorer of LSTM-DQN to assign probability scores to the possible actions.
\citet{yuan2018counting} used LSTM units in the action scorer of LSTM-DRQN to handle the better generalization.
\citet{chaudhury-etal-2020-bootstrapped} further improved generalization and reduced overfitting by training a bootstrapped model, named CREST, on context-relevant observation text.
\citet{adolphs2019ledeepchef} presented one of the winning strategies in the First-TextWorld Competition using the actor-critic algorithm~\cite{mnih2016asynchronous}  for training the policy. Unlike these text-only models, NESTA uses symbolic reasoning over the lifted rules for better generalization and interpretability.

\textbf{Graph-based Agents:}
Instead of relying on the neural models to capture the structure of the observed text %and dependencies across the game interactions
, recent works considered the graph representation of the observed text to guide the agent for better exploration. Graph-based agents from \cite{ammanabrolu2018playing, ammanabrolu2020graph} build a knowledge graph representation of the textual state for efficient exploration and handling large action space. \citet{adhikari2020learning} learns a dynamic belief graph from raw observations using adversarial learning on the First Textworld Problems (FTWP). \citet{atzeni2021case} proposed a case-based reasoning approach that improves upon existing graph-based methods by reusing the past positive experiences stored in the agent's memory. Unlike NESTA, these graph-based methods suffer from noise in the observation as the graphs are generated from the observed text.

\textbf{Reasoning-based Agents:}
Both text-only and graph-based methods use only the texts observed during the game interaction.  \citet{murugesan2020textbased} introduced Textworld commonsense (TWC), text-based cleanup games that require commonsense reasoning-based knowledge about everyday household objects 
% (e.g., milk belongs in a refrigerator, dirty towel in a laundry hamper, etc) enforcing the need for external knowledge for better exploration.  
Recent works tried to enrich text-only agents with commonsense reasoning for exploiting readily-available external knowledge graphs \cite{murugesan2021efficient} and images generated from the observed texts using pre-trained models \cite{murugesan2022eye}. These methods suffer from noisy features extracted from the external knowledge thus hindering the learning ability of the text-based RL agents. Unlike the traditional deep RL agents, \citet{chaudhury2021neuro, kimura2021neuro, basu2021hybrid}
proposed neuro-symbolic agents for TBGs that show near-perfect performance. Related work from \citet{li2021implicit} uses the world model as a symbolic representation to capture the current state of the game.  These approaches require hand-engineering of domain-specific symbolic state representation. On the other hand, NESTA presents a generic domain-independent symbolic logic representation with an automatic symbolic rule learner that handles large action spaces and noisy observation with ease. 

In other symbolic methods, there are works~\cite{petersendeep,costa2020fast} which employ deep learning for neuro-symbolic regression.
Compared to these methods, NESTA aims to improve the generalization to unseen cases, whereas these methods train and test in the same setting. Additionally, neuro-symbolic regression methods have limited interaction with the environment in intermediate steps, and reward is obtained at the terminal state. However, for NESTA we use the symbolic representation from intermediate steps to learn action rules from partially-observable symbolic states.

%% file: sections/conclusion.tex
\section{Conclusion}
In this paper, we present NESTA, a neuro-symbolic policy learning method that modularizes language understanding using an AMR-based semantic parsing module and RL policy optimization using an ILP rule learner. NESTA benefits from prior advances in AMR-based generic parsers for symbolic fact extraction allowing the ILP symbolic learner to solely learn interpretable action rules. NESTA outperforms SOTA models on TBGs by showing better generalization while learning from a fewer number of training interactions. We believe our model is one of the first works combining advances in neural semantic parsing and efficient symbolic planning for text-based RL. We hope this work will encourage future research in this direction.

%% file: acl2023_NESTA_arxiv.bbl
\begin{thebibliography}{42}
\expandafter\ifx\csname natexlab\endcsname\relax\def\natexlab#1{#1}\fi

\bibitem[{Adhikari et~al.(2020)Adhikari, Yuan, C\^{o}t\'{e}, Zelinka, Rondeau,
  Laroche, Poupart, Tang, Trischler, and Hamilton}]{adhikari2020learning}
Ashutosh Adhikari, Xingdi Yuan, Marc-Alexandre C\^{o}t\'{e}, Mikul\'{a}\v{s}
  Zelinka, Marc-Antoine Rondeau, Romain Laroche, Pascal Poupart, Jian Tang,
  Adam Trischler, and Will Hamilton. 2020.
\newblock \href
  {https://proceedings.neurips.cc/paper/2020/file/1fc30b9d4319760b04fab735fbfed9a9-Paper.pdf}
  {Learning dynamic belief graphs to generalize on text-based games}.
\newblock In \emph{Advances in Neural Information Processing Systems},
  volume~33, pages 3045--3057. Curran Associates, Inc.

\bibitem[{Adolphs and Hofmann(2020)}]{adolphs2019ledeepchef}
Leonard Adolphs and Thomas Hofmann. 2020.
\newblock \href {https://doi.org/10.1609/aaai.v34i05.6228} {Ledeepchef deep
  reinforcement learning agent for families of text-based games}.
\newblock \emph{Proceedings of the AAAI Conference on Artificial Intelligence},
  34(05):7342--7349.

\bibitem[{Ammanabrolu and Hausknecht(2020)}]{ammanabrolu2020graph}
Prithviraj Ammanabrolu and Matthew Hausknecht. 2020.
\newblock \href {https://openreview.net/forum?id=B1x6w0EtwH} {Graph constrained
  reinforcement learning for natural language action spaces}.
\newblock In \emph{International Conference on Learning Representations}.

\bibitem[{Ammanabrolu and Riedl(2019)}]{ammanabrolu2018playing}
Prithviraj Ammanabrolu and Mark Riedl. 2019.
\newblock \href {https://doi.org/10.18653/v1/N19-1358} {Playing text-adventure
  games with graph-based deep reinforcement learning}.
\newblock In \emph{Proceedings of the 2019 Conference of the North {A}merican
  Chapter of the Association for Computational Linguistics: Human Language
  Technologies, Volume 1 (Long and Short Papers)}, pages 3557--3565,
  Minneapolis, Minnesota. Association for Computational Linguistics.

\bibitem[{Atzeni et~al.(2021)Atzeni, Dhuliawala, Murugesan, and
  Sachan}]{atzeni2021case}
Mattia Atzeni, Shehzaad~Zuzar Dhuliawala, Keerthiram Murugesan, and Mrinmaya
  Sachan. 2021.
\newblock Case-based reasoning for better generalization in textual
  reinforcement learning.
\newblock In \emph{International Conference on Learning Representations}.

\bibitem[{Banarescu et~al.(2013)Banarescu, Bonial, Cai, Georgescu, Griffitt,
  Hermjakob, Knight, Koehn, Palmer, and Schneider}]{banarescu2013abstract}
Laura Banarescu, Claire Bonial, Shu Cai, Madalina Georgescu, Kira Griffitt, Ulf
  Hermjakob, Kevin Knight, Philipp Koehn, Martha Palmer, and Nathan Schneider.
  2013.
\newblock Abstract meaning representation for sembanking.
\newblock In \emph{Proceedings of the 7th linguistic annotation workshop and
  interoperability with discourse}, pages 178--186.

\bibitem[{Basu et~al.(2021)Basu, Murugesan, Atzeni, Kapanipathi, Talamadupula,
  Klinger, Campbell, Sachan, and Gupta}]{basu2021hybrid}
Kinjal Basu, Keerthiram Murugesan, Mattia Atzeni, Pavan Kapanipathi, Kartik
  Talamadupula, Tim Klinger, Murray Campbell, Mrinmaya Sachan, and Gopal Gupta.
  2021.
\newblock A hybrid neuro-symbolic approach for text-based games using inductive
  logic programming.
\newblock In \emph{Combining Learning and Reasoning: Programming Languages,
  Formalisms, and Representations}.

\bibitem[{Cai and Knight(2013)}]{cai2013smatch}
Shu Cai and Kevin Knight. 2013.
\newblock Smatch: an evaluation metric for semantic feature structures.
\newblock In \emph{Proceedings of the 51st Annual Meeting of the Association
  for Computational Linguistics (Volume 2: Short Papers)}, pages 748--752.

\bibitem[{Chaudhury et~al.(2020)Chaudhury, Kimura, Talamadupula, Tatsubori,
  Munawar, and Tachibana}]{chaudhury-etal-2020-bootstrapped}
Subhajit Chaudhury, Daiki Kimura, Kartik Talamadupula, Michiaki Tatsubori, Asim
  Munawar, and Ryuki Tachibana. 2020.
\newblock Bootstrapped {Q}-learning with context relevant observation pruning
  to generalize in text-based games.
\newblock In \emph{Proceedings of the 2020 Conference on Empirical Methods in
  Natural Language Processing (EMNLP)}, pages 3002--3008.

\bibitem[{Chaudhury et~al.(2021)Chaudhury, Sen, Ono, Kimura, Tatsubori, and
  Munawar}]{chaudhury2021neuro}
Subhajit Chaudhury, Prithviraj Sen, Masaki Ono, Daiki Kimura, Michiaki
  Tatsubori, and Asim Munawar. 2021.
\newblock Neuro-symbolic approaches for text-based policy learning.
\newblock In \emph{Proceedings of the 2021 Conference on Empirical Methods in
  Natural Language Processing}, pages 3073--3078.

\bibitem[{Costa et~al.(2020)Costa, Dangovski, Dugan, Kim, Goyal,
  Solja{\v{c}}i{\'c}, and Jacobson}]{costa2020fast}
Allan Costa, Rumen Dangovski, Owen Dugan, Samuel Kim, Pawan Goyal, Marin
  Solja{\v{c}}i{\'c}, and Joseph Jacobson. 2020.
\newblock Fast neural models for symbolic regression at scale.
\newblock \emph{arXiv preprint arXiv:2007.10784}.

\bibitem[{C{\^o}t{\'e} et~al.(2018)C{\^o}t{\'e}, K{\'a}d{\'a}r, Yuan, Kybartas,
  Barnes, Fine, Moore, Hausknecht, Asri, Adada et~al.}]{cote2018textworld}
Marc-Alexandre C{\^o}t{\'e}, {\'A}kos K{\'a}d{\'a}r, Xingdi Yuan, Ben Kybartas,
  Tavian Barnes, Emery Fine, James Moore, Matthew Hausknecht, Layla~El Asri,
  Mahmoud Adada, et~al. 2018.
\newblock Textworld: A learning environment for text-based games.
\newblock \emph{arXiv preprint arXiv:1806.11532}.

\bibitem[{Dong et~al.(2019)Dong, Mao, Lin, Wang, Li, and Zhou}]{dong2019neural}
Honghua Dong, Jiayuan Mao, Tian Lin, Chong Wang, Lihong Li, and Denny Zhou.
  2019.
\newblock Neural logic machines.
\newblock \emph{arXiv preprint arXiv:1904.11694}.

\bibitem[{Galitsky(2020)}]{galitsky2020employing}
Boris Galitsky. 2020.
\newblock Employing abstract meaning representation to lay the last-mile toward
  reading comprehension.
\newblock In \emph{Artificial Intelligence for Customer Relationship
  Management}, pages 57--86. Springer.

\bibitem[{Hausknecht et~al.(2020)Hausknecht, Ammanabrolu, C{\^o}t{\'e}, and
  Yuan}]{hausknecht2020interactive}
Matthew Hausknecht, Prithviraj Ammanabrolu, Marc-Alexandre C{\^o}t{\'e}, and
  Xingdi Yuan. 2020.
\newblock Interactive fiction games: A colossal adventure.
\newblock In \emph{Proceedings of the AAAI Conference on Artificial
  Intelligence}, volume~34, pages 7903--7910.

\bibitem[{He et~al.(2016)He, Chen, He, Gao, Li, Deng, and
  Ostendorf}]{he-etal-2016-deep-reinforcement}
Ji~He, Jianshu Chen, Xiaodong He, Jianfeng Gao, Lihong Li, Li~Deng, and Mari
  Ostendorf. 2016.
\newblock \href {https://doi.org/10.18653/v1/P16-1153} {Deep reinforcement
  learning with a natural language action space}.
\newblock In \emph{Proceedings of the 54th Annual Meeting of the Association
  for Computational Linguistics (Volume 1: Long Papers)}, pages 1621--1630.

\bibitem[{Jiang and Luo(2019)}]{jiang2019neural}
Zhengyao Jiang and Shan Luo. 2019.
\newblock Neural logic reinforcement learning.
\newblock In \emph{International conference on machine learning}, pages
  3110--3119. PMLR.

\bibitem[{Kaelbling et~al.(1998)Kaelbling, Littman, and
  Cassandra}]{kaelbling1998planning}
Leslie~Pack Kaelbling, Michael~L Littman, and Anthony~R Cassandra. 1998.
\newblock Planning and acting in partially observable stochastic domains.
\newblock \emph{Artificial intelligence}, 101(1-2):99--134.

\bibitem[{Kapanipathi et~al.(2021)Kapanipathi, Abdelaziz, Ravishankar, Roukos,
  Gray, Astudillo, Chang, Cornelio, Dana, Fokoue-Nkoutche
  et~al.}]{kapanipathi2021leveraging}
Pavan Kapanipathi, Ibrahim Abdelaziz, Srinivas Ravishankar, Salim Roukos,
  Alexander Gray, Ram{\'o}n~Fernandez Astudillo, Maria Chang, Cristina
  Cornelio, Saswati Dana, Achille Fokoue-Nkoutche, et~al. 2021.
\newblock Leveraging abstract meaning representation for knowledge base
  question answering.
\newblock In \emph{Findings of the Association for Computational Linguistics:
  ACL-IJCNLP 2021}, pages 3884--3894.

\bibitem[{Kimura et~al.(2021)Kimura, Ono, Chaudhury, Kohita, Wachi, Agravante,
  Tatsubori, Munawar, and Gray}]{kimura2021neuro}
Daiki Kimura, Masaki Ono, Subhajit Chaudhury, Ryosuke Kohita, Akifumi Wachi,
  Don~Joven Agravante, Michiaki Tatsubori, Asim Munawar, and Alexander Gray.
  2021.
\newblock Neuro-symbolic reinforcement learning with first-order logic.
\newblock In \emph{Proceedings of the 2021 Conference on Empirical Methods in
  Natural Language Processing}, pages 3505--3511.

\bibitem[{Kingsbury and Palmer(2002)}]{kingsbury2002treebank}
Paul~R Kingsbury and Martha Palmer. 2002.
\newblock From treebank to propbank.
\newblock In \emph{LREC}, pages 1989--1993. Citeseer.

\bibitem[{Lee et~al.(2022)Lee, Astudillo, Thanh~Lam, Naseem, Florian, and
  Roukos}]{lee-etal-2022-maximum}
Young-Suk Lee, Ram{\'o}n Astudillo, Hoang Thanh~Lam, Tahira Naseem, Radu
  Florian, and Salim Roukos. 2022.
\newblock \href {https://doi.org/10.18653/v1/2022.naacl-main.393} {Maximum
  {B}ayes {S}match ensemble distillation for {AMR} parsing}.
\newblock In \emph{Proceedings of the 2022 Conference of the North American
  Chapter of the Association for Computational Linguistics: Human Language
  Technologies}, pages 5379--5392. Association for Computational Linguistics.

\bibitem[{Li et~al.(2021)Li, Nye, and Andreas}]{li2021implicit}
Belinda~Z Li, Maxwell Nye, and Jacob Andreas. 2021.
\newblock Implicit representations of meaning in neural language models.
\newblock In \emph{Proceedings of the 59th Annual Meeting of the Association
  for Computational Linguistics and the 11th International Joint Conference on
  Natural Language Processing (Volume 1: Long Papers)}, pages 1813--1827.

\bibitem[{Lien and Kouylekov(2015)}]{lien2015semantic}
Elisabeth Lien and Milen Kouylekov. 2015.
\newblock Semantic parsing for textual entailment.
\newblock In \emph{Proceedings of the 14th International Conference on Parsing
  Technologies}, pages 40--49.

\bibitem[{Lu et~al.(2021)Lu, West, Zellers, Le~Bras, Bhagavatula, and
  Choi}]{lu2021neurologic}
Ximing Lu, Peter West, Rowan Zellers, Ronan Le~Bras, Chandra Bhagavatula, and
  Yejin Choi. 2021.
\newblock Neurologic decoding:(un) supervised neural text generation with
  predicate logic constraints.
\newblock In \emph{NAACL-HLT}.

\bibitem[{Ma et~al.(2021)Ma, Zhuang, Weng, Zhuo, Li, Liu, and
  Hao}]{ma2021learning}
Zhihao Ma, Yuzheng Zhuang, Paul Weng, Hankz~Hankui Zhuo, Dong Li, Wulong Liu,
  and Jianye Hao. 2021.
\newblock Learning symbolic rules for interpretable deep reinforcement
  learning.
\newblock \emph{arXiv preprint arXiv:2103.08228}.

\bibitem[{Mitra and Baral(2016)}]{mitra2016addressing}
Arindam Mitra and Chitta Baral. 2016.
\newblock Addressing a question answering challenge by combining statistical
  methods with inductive rule learning and reasoning.
\newblock In \emph{Proceedings of the AAAI Conference on Artificial
  Intelligence}, volume~30.

\bibitem[{Mnih et~al.(2016)Mnih, Badia, Mirza, Graves, Lillicrap, Harley,
  Silver, and Kavukcuoglu}]{mnih2016asynchronous}
Volodymyr Mnih, Adria~Puigdomenech Badia, Mehdi Mirza, Alex Graves, Timothy
  Lillicrap, Tim Harley, David Silver, and Koray Kavukcuoglu. 2016.
\newblock Asynchronous methods for deep reinforcement learning.
\newblock In \emph{International conference on machine learning}, pages
  1928--1937. PMLR.

\bibitem[{Murugesan et~al.(2020)Murugesan, Atzeni, Kapanipathi, Shukla,
  Kumaravel, Tesauro, Talamadupula, Sachan, and
  Campbell}]{murugesan2020textbased}
Keerthiram Murugesan, Mattia Atzeni, Pavan Kapanipathi, Pushkar Shukla, Sadhana
  Kumaravel, Gerald Tesauro, Kartik Talamadupula, Mrinmaya Sachan, and Murray
  Campbell. 2020.
\newblock \href {http://arxiv.org/abs/2010.03790} {Text-based rl agents with
  commonsense knowledge: New challenges, environments and baselines}.

\bibitem[{Murugesan et~al.(2021)Murugesan, Atzeni, Kapanipathi, Talamadupula,
  Sachan, and Campbell}]{murugesan2021efficient}
Keerthiram Murugesan, Mattia Atzeni, Pavan Kapanipathi, Kartik Talamadupula,
  Mrinmaya Sachan, and Murray Campbell. 2021.
\newblock Efficient text-based reinforcement learning by jointly leveraging
  state and commonsense graph representations.
\newblock In \emph{Proceedings of the 59th Annual Meeting of the Association
  for Computational Linguistics and the 11th International Joint Conference on
  Natural Language Processing (Volume 2: Short Papers)}, pages 719--725.

\bibitem[{Murugesan et~al.(2022)Murugesan, Chaudhury, and
  Talamadupula}]{murugesan2022eye}
Keerthiram Murugesan, Subhajit Chaudhury, and Kartik Talamadupula. 2022.
\newblock Eye of the beholder: Improved relation generalization for text-based
  reinforcement learning agents.
\newblock In \emph{Proceedings of the AAAI Conference on Artificial
  Intelligence}, volume~36, pages 11094--11102.

\bibitem[{Narasimhan et~al.(2015{\natexlab{a}})Narasimhan, Kulkarni, and
  Barzilay}]{narasimhan-etal-2015-language}
Karthik Narasimhan, Tejas Kulkarni, and Regina Barzilay. 2015{\natexlab{a}}.
\newblock \href {https://doi.org/10.18653/v1/D15-1001} {Language understanding
  for text-based games using deep reinforcement learning}.
\newblock In \emph{Proceedings of the 2015 Conference on Empirical Methods in
  Natural Language Processing}, pages 1--11.

\bibitem[{Narasimhan et~al.(2015{\natexlab{b}})Narasimhan, Kulkarni, and
  Barzilay}]{narasimhan2015language}
Karthik Narasimhan, Tejas Kulkarni, and Regina Barzilay. 2015{\natexlab{b}}.
\newblock Language understanding for text-based games using deep reinforcement
  learning.
\newblock In \emph{Proceedings of the 2015 Conference on Empirical Methods in
  Natural Language Processing}, pages 1--11.

\bibitem[{Petersen et~al.()Petersen, Larma, Mundhenk, Santiago, Kim, and
  Kim}]{petersendeep}
Brenden~K Petersen, Mikel~Landajuela Larma, Terrell~N Mundhenk, Claudio~Prata
  Santiago, Soo~Kyung Kim, and Joanne~Taery Kim.
\newblock Deep symbolic regression: Recovering mathematical expressions from
  data via risk-seeking policy gradients.
\newblock In \emph{International Conference on Learning Representations}.

\bibitem[{Riegel et~al.(2020)Riegel, Gray, Luus, Khan, Makondo, Akhalwaya,
  Qian, Fagin, Barahona, Sharma et~al.}]{riegel2020logical}
Ryan Riegel, Alexander Gray, Francois Luus, Naweed Khan, Ndivhuwo Makondo,
  Ismail~Yunus Akhalwaya, Haifeng Qian, Ronald Fagin, Francisco Barahona, Udit
  Sharma, et~al. 2020.
\newblock Logical neural networks.
\newblock \emph{arXiv preprint arXiv:2006.13155}.

\bibitem[{Sen et~al.(2022)Sen, de~Carvalho, Riegel, and Gray}]{sen2022neuro}
Prithviraj Sen, Breno~WSR de~Carvalho, Ryan Riegel, and Alexander Gray. 2022.
\newblock Neuro-symbolic inductive logic programming with logical neural
  networks.
\newblock In \emph{Proceedings of the AAAI Conference on Artificial
  Intelligence}, volume~36, pages 8212--8219.

\bibitem[{Speer et~al.(2017)Speer, Chin, and Havasi}]{speer2017conceptnet}
Robyn Speer, Joshua Chin, and Catherine Havasi. 2017.
\newblock Conceptnet 5.5: An open multilingual graph of general knowledge.
\newblock In \emph{Thirty-first AAAI conference on artificial intelligence}.

\bibitem[{Van~Noord and Bos(2017)}]{van2017neural}
Rik Van~Noord and Johan Bos. 2017.
\newblock Neural semantic parsing by character-based translation: Experiments
  with abstract meaning representations.
\newblock \emph{arXiv preprint arXiv:1705.09980}.

\bibitem[{Watkins and Dayan(1992)}]{watkins1992q}
Christopher~JCH Watkins and Peter Dayan. 1992.
\newblock Q-learning.
\newblock \emph{Machine learning}, 8(3-4):279--292.

\bibitem[{Yuan et~al.(2018)Yuan, C{\^o}t{\'e}, Sordoni, Laroche, Combes,
  Hausknecht, and Trischler}]{yuan2018counting}
Xingdi Yuan, Marc-Alexandre C{\^o}t{\'e}, Alessandro Sordoni, Romain Laroche,
  Remi Tachet~des Combes, Matthew Hausknecht, and Adam Trischler. 2018.
\newblock Counting to explore and generalize in text-based games.
\newblock \emph{arXiv preprint arXiv:1806.11525}.

\bibitem[{Yuan et~al.(2019)Yuan, Côté, Sordoni, Laroche, des Combes,
  Hausknecht, and Trischler}]{yuan2019counting}
Xingdi Yuan, Marc-Alexandre Côté, Alessandro Sordoni, Romain Laroche,
  Remi~Tachet des Combes, Matthew Hausknecht, and Adam Trischler. 2019.
\newblock \href {http://arxiv.org/abs/1806.11525} {Counting to explore and
  generalize in text-based games}.

\bibitem[{Zhou et~al.(2021)Zhou, Naseem, Astudillo, and Florian}]{zhou2021amr}
Jiawei Zhou, Tahira Naseem, Ram{\'o}n~Fernandez Astudillo, and Radu Florian.
  2021.
\newblock Amr parsing with action-pointer transformer.
\newblock \emph{arXiv preprint arXiv:2104.14674}.

\end{thebibliography}
